\documentclass[conference]{IEEEtran}
\usepackage{cite}
\usepackage{amsmath,amssymb,amsfonts}
\usepackage{algorithmic}
\usepackage{graphicx}
\usepackage{textcomp}
\usepackage{xcolor}
\usepackage{array}
\usepackage{booktabs}

\def\BibTeX{{\rm B\kern-.05em{\sc i\kern-.025em b}\kern-.08em
    T\kern-.1667em\lower.7ex\hbox{E}\kern-.125emX}}



\begin{document}

\title{ProsMAE: Multi-Source MAE Pretraining for ISUP Grade Classification\\
}

\author{%

\makebox[\textwidth][c]{%
\resizebox{\textwidth}{!}{%
\begin{tabular}
{@{\extracolsep{\fill}}ccccc@{}}
Anna Jung &
Kyeonghun Kim &
Youngung Han &
Eunseob Choi &
Jiwon Yang \\[0.1em]
\textit{SNU} &
\textit{OUTTA} &
\textit{SNU} &
\textit{GIST} &
\textit{SNU} \\[0.1em]
{\fontsize{8.8}{8.8}\selectfont annajung227@snu.ac.kr} &
{\fontsize{8.8}{8.8}\selectfont kyeonghun.kim@outta.ai} &
{\fontsize{8.8}{8.8}\selectfont yuhan@snu.ac.kr} &
{\fontsize{8.8}{8.8}\selectfont eunseobchoi@gm.gist.ac.kr} &
{\fontsize{8.8}{8.8}\selectfont jwyang29@snu.ac.kr}
\end{tabular}%
}%
}
\\[1.5ex]

\makebox[\textwidth][c]{%
\resizebox{\textwidth}{!}{%
\begin{tabular}
{@{\extracolsep{\fill}}ccccc@{}}
Yunjin Seo &
Ken Ying-Kai Liao &
Pa Hong &
Hyuk-Jae Lee &
Nam-Joon Kim\textsuperscript{\dag} \\[0.2em]
\textit{SNU} &
\textit{NVIDIA} &
\textit{Samsung Changwon Hospital} &
\textit{SNU} &
\textit{SNU} \\[0.1em]
{\fontsize{8.8}{8.8}\selectfont tjdbswls1456@snu.ac.kr} &
{\fontsize{8.8}{8.8}\selectfont kenyingkail@nvidia.com} &
{\fontsize{8.8}{8.8}\selectfont papa.hong@samsung.com} &
{\fontsize{8.8}{8.8}\selectfont hjlee@capp.snu.ac.kr} &
{\fontsize{8.8}{8.8}\selectfont knj01@snu.ac.kr}
\end{tabular}%
}%
}

}
\maketitle
\begingroup
\renewcommand\thefootnote{}
\footnotetext{%
\noindent\rule{1.6in}{0.4pt}\\[0.5ex]
\textsuperscript{\dag} Corresponding author.\\
SNU and GIST stand for Seoul National University and Gwangju Institute of Science and Technology, respectively.
}
\endgroup

\begin{abstract}
Whole slide images (WSIs) provide rich diagnostic information for computational pathology, but their gigapixel scale, stain variation, scanner differences, tissue artifacts, and limited expert annotation make robust model training challenging. This paper presents a multi-source Masked Autoencoder (MAE) framework, named ProsMAE, for histopathology representation learning. Tiles from Prostate cANcer graDe Assessment (PANDA), CAncer MEtastases in LYmph nOdes challeNge 2017 (CAMELYON17), and BReAst Carcinoma Subtyping (BRACS) are used for ProsMAE pretraining to expose the encoder to diverse tissue morphology and acquisition conditions. The learned encoder is transferred for International Society of Urological Pathology (ISUP) grade classification through ProsCLS, using a frozen encoder and a linear classification head. ProsMAE achieved a higher mean validation quadratic weighted kappa (QWK) than the vanilla MAE frozen linear-probe baseline under the evaluated disjoint PANDA split. Repeated-split evaluation remains necessary to establish robustness across split compositions. 
\end{abstract}

\begin{IEEEkeywords}
Whole slide image analysis, self-supervised representation learning, Gleason grading, prostate cancer diagnosis, digital pathology
\end{IEEEkeywords}

\section{Introduction}

Whole slide images (WSIs) are central to computational pathology because they preserve tissue morphology at high resolution. However, their gigapixel scale makes direct processing computationally difficult, so most pipelines rely on tile-based analysis and slide-level aggregation to connect local tissue patterns with diagnostic labels \cite{xu2024provgigapath,campanella2019clinical,lu2021data}. In prostate cancer, this is important because the International Society of Urological
Pathology (ISUP) grading depends on glandular architecture and morphological patterns that may appear only in limited biopsy regions \cite{bulten2022artificial}.

Supervised WSI learning is limited by expensive expert annotation and weak alignment between slide-level labels and local tissue morphology \cite{campanella2019clinical,lu2021data}. Self-supervised learning helps address this by learning transferable representations from unlabeled pathology tiles before downstream classification \cite{campanella2025benchmark,chen2026beyondvit}. Among these methods, Masked Autoencoders (MAEs) are particularly suitable because they reconstruct missing image patches from visible tissue context using Vision Transformer (ViT) patch tokens \cite{he2022masked,dosovitskiy2021image}. Recent work has also demonstrated the effectiveness of MAE in medical imaging \cite{kim2026maesil}.

A remaining challenge is domain variation caused by differences in scanners, staining protocols, and tissue preparation~\cite{aubreville2022midog,tellez2019quantifying}. To improve robustness, we propose ProsMAE, a multi-source MAE pretraining framework. By pretraining on Prostate cANcer graDe Assessment (PANDA), CAncer MEtastases in LYmph nOdes challeNge 2017 (CAMELYON17), and BReAst Carcinoma Subtyping (BRACS) with a high mask ratio, the encoder learns morphology-preserving features that are less sensitive to dataset-specific variation \cite{bulten2022artificial,bandi2018detection,brancati2022bracs}.

The main contributions are summarized as follows:
\begin{itemize}
\item We propose ProsMAE, a multi-source MAE pretraining framework for WSI representation learning.
\item We use PANDA, BRACS, and CAMELYON17 for diverse histopathology pretraining.
\item We transfer the encoder to frozen linear-probe ISUP grade classification through ProsCLS.
\end{itemize}

\begin{figure*}[t]
\centering
\includegraphics[width=\textwidth]{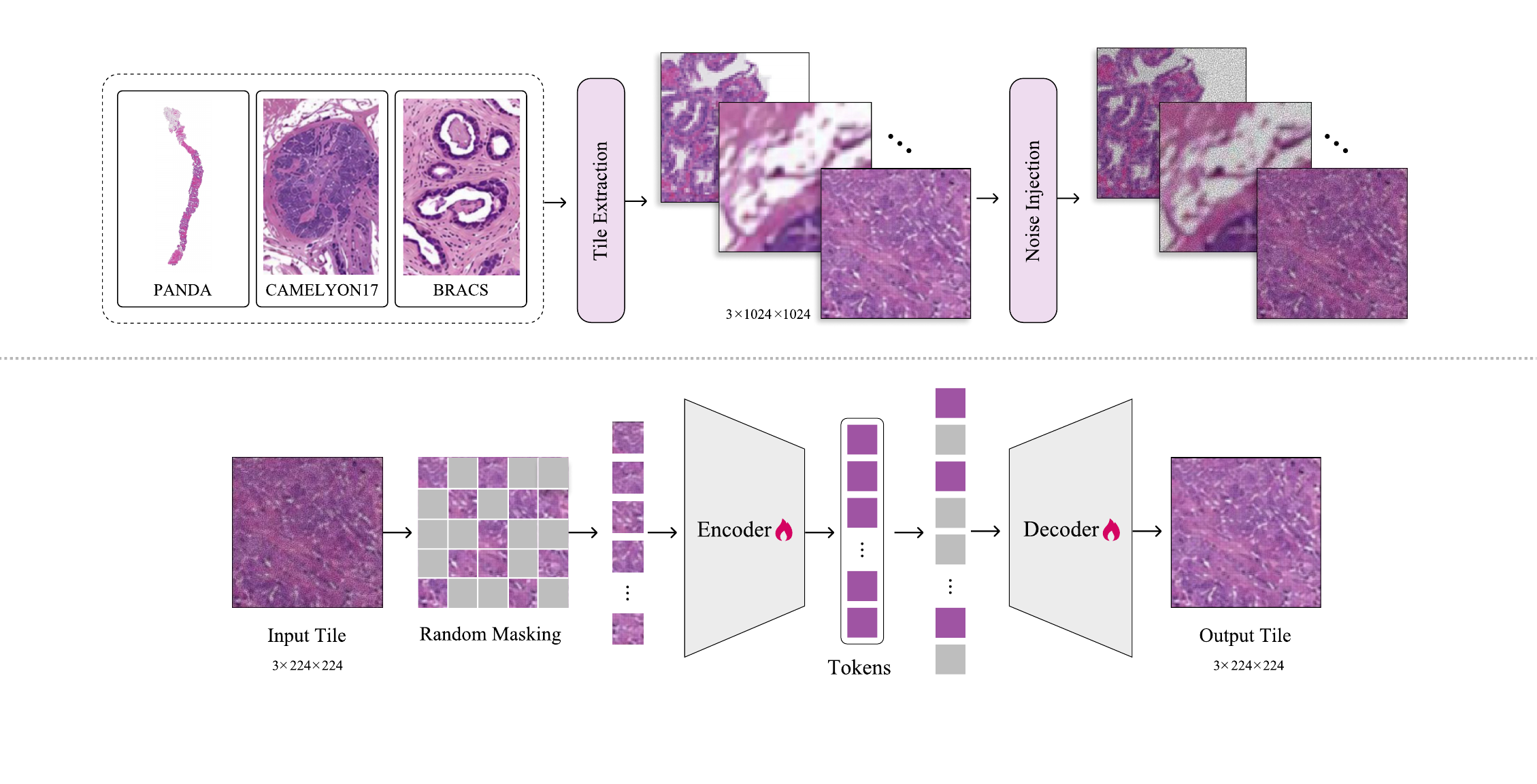}
\caption{ProsMAE pretraining workflow. WSI tiles from PANDA, CAMELYON17, and BRACS are randomly masked, encoded, and reconstructed to learn multi-source histopathology representations.}
\label{fig:mae_pretraining}
\end{figure*}

\section{Methodology}

The proposed framework consists of two stages: ProsMAE, the multi-source MAE pretraining stage, and ProsCLS, the downstream ISUP grade classification stage.

\begin{table}[htbp]
\caption{External Data Source Results}
\centering
\small
\renewcommand{\arraystretch}{1.1}
\setlength{\tabcolsep}{4pt}

\begin{tabular}{lcc}
\toprule
\textbf{Setting} & \textbf{Best QWK} & \textbf{Std} \\
\midrule
PANDA-only & 0.3757 & 0.0204 \\
PANDA{+}CAMELYON17 & 0.4165 & 0.0339 \\
PANDA{+}BRACS & 0.4330 & 0.0584 \\
\midrule
\textbf{PANDA{+}CAMELYON17{+}BRACS} & \textbf{0.4734} & \textbf{0.0104} \\
\bottomrule
\end{tabular}

\label{tab:source_methodology}
\end{table}

As shown in Table~\ref{tab:source_methodology}, three public histopathology datasets are used for representation learning: PANDA for prostate cancer, CAMELYON17 for lymph node metastasis, and BRACS for breast cancer subtype. PANDA also serves as our downstream evaluation cohort. The downstream task is formulated as a six-class classification, where Class 0 denotes benign/no-cancer biopsies and Classes 1-5 denote ISUP grade groups 1-5.

\subsection{Masked Autoencoder Backbone}

In the first stage, a pretrained MAE is adapted to unlabeled tiles from these three datasets via masked image reconstruction, as shown in Fig.~\ref{fig:mae_pretraining}. Following the standard MAE design \cite{he2022masked}, the masking ratio is set to 0.75. ViTs divide each tile into fixed-size patches and process the resulting patch embeddings as a token sequence \cite{dosovitskiy2021image}. The encoder processes only visible patch tokens, while the decoder reconstructs masked regions from latent representations and mask tokens.

\subsection{Noise Injection Ablation}

Gaussian noise is added to the input before masked reconstruction, while the target remains the original clean image. This ablation tests whether reconstructing from corrupted inputs improves the robustness of representation against typical clinical variations, such as differences in scanners, blur, and stain variability \cite{tellez2019quantifying,Wangetal2024}.
Given an input tile $x$, a Gaussian noise transformation $\mathcal{N}_{\sigma}(\cdot)$ produces:
\begin{equation}
\tilde{x} = \mathcal{N}_{\sigma}(x),
\end{equation}
where $\sigma$ denotes the noise standard deviation. We evaluate noise levels $\sigma \in \{0.02, 0.05, 0.10, 0.20\}$. The noisy tile $\tilde{x}$ is divided into non-overlapping patches, randomly masked, and reconstructed.

\begin{figure*}[htbp]
\centerline{\includegraphics[width=\textwidth]{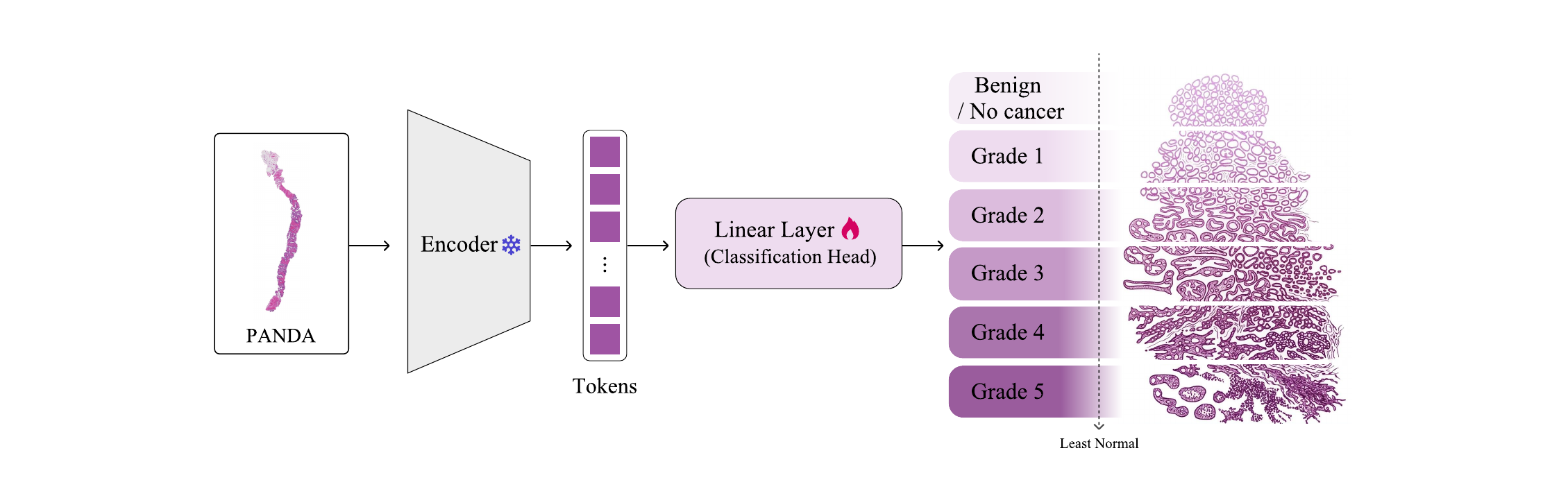}}
\caption{ProsCLS downstream classification workflow. PANDA WSIs are divided into tiles, features are extracted using the frozen ProsMAE encoder, tile-level features are aggregated by mean pooling, and a linear classification head predicts the downstream six-class label (Benign/No cancer + ISUP Grades 1-5).}
\label{fig:classification}
\end{figure*}

The reconstruction target is the original normalized image patch values. The MAE reconstruction loss is computed over the masked patches:

\begin{equation}
\mathcal{L}_{MAE} =
\frac{1}{|\Omega|}
\sum_{i \in \Omega}
\|x_i - \hat{x}_i\|_2^2,
\end{equation}

where $\Omega$ is the set of masked patches, $x_i$ is the target patch, and $\hat{x}_i$ is the reconstructed patch.

\subsection{ISUP Grade Classification}

After MAE pretraining, the decoder is removed and the encoder is transferred to ProsCLS for PANDA ISUP grade classification as shown in Fig.~\ref{fig:classification}. This follows the standard linear evaluation setting in self-supervised learning, where a lightweight classifier is trained on learned representations to assess feature quality \cite{chen2020simple,ciga2022self,he2022masked}.

For a slide containing $N$ sampled tiles $\{x_1, x_2, ..., x_N\}$, the encoder extracts a feature vector from each tile:

\begin{equation}
z_i = f_{\theta}(x_i), \quad i = 1,2,\dots,N
\end{equation}

where $f_{\theta}$ denotes the pretrained MAE encoder.

The tile-level features are aggregated into a slide-level representation using mean pooling:

\begin{equation}
z_{slide} =
\frac{1}{N}
\sum_{i=1}^{N} z_i
\end{equation}

A linear classification head predicts the ISUP grade label:

\begin{equation}
\hat{y} =
\text{Softmax}(Wz_{slide} + b).
\end{equation}

The classifier is trained using cross-entropy loss:

\begin{equation}
\mathcal{L}_{cls} =
-\sum_{c=1}^{C} y_c \log(\hat{y}_c),
\end{equation}

where $C = 6$ is the number of downstream classes, including ISUP Grades 1-5 and the additional benign/no-cancer class.

\section{Experiments}

\subsection{Experimental Setup}

We evaluate whether multi-source MAE pretraining improves downstream ISUP classification. The backbone is initialized with Facebook/Meta ViT-MAE-Base weights (ViT-B/16, pretrained on ImageNet-1K) \cite{he2022masked}. ProsMAE pretraining uses a mask ratio of 0.75 without additional noise.

To prevent data leakage, PANDA slides are divided into disjoint sets: 241 for pretraining, 82 for downstream training, and 80 for validation. WSIs are partitioned into $1024 {\times} 1024$ tissue regions from Level 2 images ($\approx 8.0~\mu\mathrm{m}$/pixel). For MAE pretraining, these regions are resized to $224 {\times} 224$. For downstream evaluation, $512 {\times} 512$ tiles are extracted and resized to $224 {\times} 224$ before feature extraction.

Stain normalization is omitted to preserve natural stain variation and encourage robustness to heterogeneous acquisition conditions \cite{tellez2019quantifying,aubreville2022midog}. Pretraining uses AdamW for up to 5000 steps, capped at 20 epochs, with a batch size of 64, a learning rate of $5{\times}10^{-5}$, a 250-step warmup and cosine decay. Each WSI contributes 100 tiles. During evaluation, the encoder is frozen and a linear classifier with balanced class weights is trained on mean-pooled features of 100 tiles per slide.



\subsection{Evaluation Metrics}

Performance is evaluated using accuracy, macro F1-score, and quadratic weighted kappa (QWK) \cite{cohen1968weighted}, a weighted agreement metric that penalizes larger ordinal disagreements more strongly. Macro F1-score computes F1 for each class and averages them equally, providing a balanced evaluation across common and minority grades \cite{Takahashietal2021}. QWK is used because ISUP grades are ordinal, where errors between adjacent grades are less severe than errors between distant grades \cite{bulten2022artificial}.

\subsection{Reconstruction Performance}

We compare the reconstruction performance of ProsMAE with standard Autoencoder (AE), Variational Autoencoder (VAE), and single-source MAE baselines. Evaluation is performed on PANDA, CAMELYON17, BRACS, and the combined dataset using Learned Perceptual Image Patch Similarity (LPIPS), Structural Similarity Index Measure (SSIM), and Peak Signal-to-Noise Ratio (PSNR). Total pretraining GPU hours are also reported to assess computational efficiency.


\begin{table}[htbp]
\caption{Reconstruction Performance of ProsMAE and baseline models}
\centering
\small
\renewcommand{\arraystretch}{1.1}
\setlength{\tabcolsep}{4.5pt}
\begin{tabular}{llccc}
\toprule
\textbf{Dataset} & \textbf{Model} & \textbf{LPIPS $\downarrow$} & \textbf{SSIM $\uparrow$} & \textbf{PSNR $\uparrow$} \\
\midrule
PANDA & AE & 0.065 & 0.6921 & 30.058 \\
 & VAE & 0.063 & 0.7030 & 30.173 \\
 & MAE & 0.061 & 0.7290 & 30.242 \\
 & \textbf{ProsMAE} & \textbf{0.059} & \textbf{0.7430} & \textbf{31.142} \\
\midrule
CAMELYON17 & AE & 0.067 & 0.7250 & 31.391 \\
 & VAE & 0.064 & 0.7270 & 31.423 \\
 & MAE & 0.065 & 0.7310 & 31.519 \\
 & \textbf{ProsMAE} & \textbf{0.061} & \textbf{0.7330} & \textbf{31.771} \\
 \midrule
BRACS & AE & 0.059 & 0.7310 & 30.833 \\
 & VAE & 0.057 & 0.7430 & 30.821 \\
 & MAE & 0.057 & 0.7420 & 30.923 \\
 & \textbf{ProsMAE} & \textbf{0.056} & \textbf{0.7520} & \textbf{32.271} \\
\midrule
PANDA+BRACS & AE & 0.071 & 0.7220 & 29.613 \\
+CAMELYON17 & VAE & 0.069 & 0.7270 & 29.711 \\
 & MAE & 0.069 & 0.7310 & 29.687 \\
 & \textbf{ProsMAE} & \textbf{0.068} & \textbf{0.7330} & \textbf{30.006} \\
\bottomrule
\end{tabular}
\label{tab:reconstruction_performance}
\end{table}

As shown in Table~\ref{tab:reconstruction_performance}, ProsMAE achieves the best reconstruction scores among the evaluated models. On the combined dataset, ProsMAE achieves the lowest LPIPS score of 0.068, along with an SSIM of 0.7330, and a PSNR of 30.006. Furthermore, ProsMAE maintains highly competitive pretraining efficiency, requiring only 10-11 hours of GPU training time, which is faster than standard AE/VAE.

\subsection{Downstream Classification Results}

Across downstream seeds, ProsMAE achieved a higher mean QWK than vanilla MAE (0.4736 vs. 0.4084), averaged over seeds 42-51 and 42-52, respectively. These results suggest that multi-source MAE pretraining improves ordinal agreement under the current setting, despite limitations from the small validation set and variability across seeds.


\subsection{Ablation Study}

To better understand performance improvement, we conduct ablation studies on the mask ratio, noise injection, baseline split robustness, and tile sampling sensitivity.

\subsubsection{Mask Ratio}

We first compare MAE mask ratios of 0.25, 0.50, and 0.75 in Table~\ref{tab:mask_ratio} under the same disjoint split and 5000-step MAE pretraining setup.

\begin{table}[htbp]
\caption{Mask Ratio Ablation Results}
\centering
\small
\renewcommand{\arraystretch}{1.1}
\setlength{\tabcolsep}{4pt}

\begin{tabular}{ccccc}
\toprule
\textbf{Mask Ratio} & \textbf{Best QWK} & \textbf{Final QWK} & \textbf{Acc.} & \textbf{Macro-F1} \\
\midrule
0.25 & 0.3963 & 0.3781 & \textbf{0.3125} & 0.3042 \\
0.50 & 0.4274 & 0.3774 & 0.3000 & 0.2875 \\
\midrule
\textbf{0.75} & \textbf{0.4699} & \textbf{0.4656} & 0.2875 & 0.2902 \\
\bottomrule
\end{tabular}

\label{tab:mask_ratio}
\end{table}

Although the mask ratio of 0.25 achieves slightly higher accuracy, the 0.75 setting yields the highest Best QWK (0.4699) and Final QWK (0.4656). Because ISUP classification is an ordinal task, we prioritize QWK over standard accuracy to penalize larger grade discrepancies. Consequently, the 0.75 mask ratio is selected as our default pretraining configuration.

\subsubsection{Noise Injection}

\begin{table}[htbp]
\caption{Noise Injection Ablation Results\\(Multi-source, Mask ratio = 0.75)}
\centering
\small
\renewcommand{\arraystretch}{1.15}
\setlength{\tabcolsep}{6pt}

\begin{tabular}{lcc}
\toprule
\textbf{Noise} & \textbf{Best QWK} & \textbf{Std} \\
\midrule
\textbf{w/o noise} & \textbf{0.4734} & \textbf{0.0104} \\
\midrule
0.02 & 0.3694 & 0.0477 \\
0.05 & 0.3943 & 0.0513 \\
0.10 & 0.4350 & 0.0225 \\
0.20 & 0.4733 & 0.0184 \\
\bottomrule
\end{tabular}

\label{tab:noise_ablation}
\end{table}

In Table~\ref{tab:noise_ablation}, noise injection does not consistently improve downstream QWK. Lower noise levels reduce performance, while noise level 0.20 achieves performance similar to the no-noise setting. Therefore, no-noise mask 0.75 is retained as the primary result, and noise injection is treated as a supporting robustness ablation rather than the central contribution.

\subsubsection{Split Sensitivity of the Vanilla Baseline}

To evaluate whether the vanilla MAE baseline is sensitive to the PANDA train-validation split, we repeat the evaluation in three independent disjoint splits in Table~\ref{tab:split_ablation}.

\begin{table}[htbp]
\caption{Vanilla MAE Split Robustness Results}
\centering
\small
\renewcommand{\arraystretch}{1.1}
\setlength{\tabcolsep}{4pt}

\begin{tabular}{lccccc}
\toprule
\textbf{Split} & \textbf{Runs} & \textbf{Mean QWK} & \textbf{Std} & \textbf{Min} & \textbf{Max} \\

\midrule
43 & 3 & 0.2687 & 0.0286 & 0.2367 & 0.2918 \\
44 & 3 & 0.4972 & 0.0340 & 0.4616 & 0.5294 \\
45 & 3 & 0.3056 & 0.0159 & 0.2910 & 0.3225 \\
\midrule

\textbf{Overall} & \textbf{9} & \textbf{0.3572} & \textbf{0.1088} & \textbf{0.2367} & \textbf{0.5294} \\
\bottomrule
\end{tabular}

\label{tab:split_ablation}
\end{table}
As shown in Table~\ref{tab:split_ablation}, vanilla MAE performance varies across disjoint PANDA splits, indicating that ISUP classification is sensitive to split composition under the current low-compute protocol. ProsMAE improves performance on the primary disjoint split, but this should not be interpreted as evidence of consistent superiority across all splits.


\subsubsection{Tile Sampling Sensitivity}

We evaluate whether ProsMAE performance depends on selecting exactly 100 tiles per slide during feature extraction. Additional experiments are performed using 50 and 150 tiles per slide in Table~\ref{tab:tile_ablation}.

\begin{table}[htbp]
\caption{ProsMAE Tile Sampling Sensitivity Results}
\centering
\small
\renewcommand{\arraystretch}{1.1}
\setlength{\tabcolsep}{4pt}

\begin{tabular}{lccccc}
\toprule
\textbf{Tiles/Slide} & \textbf{Runs} & \textbf{Mean QWK} & \textbf{Std} & \textbf{Min} & \textbf{Max} \\
\midrule
50 & 3 & 0.5039 & 0.0203 & 0.4856 & 0.5258 \\
100 (main) & 4 & 0.4734 & 0.0104 & 0.4613 & 0.4860 \\
150 & 3 & 0.4817 & 0.0234 & 0.4607 & 0.5070 \\
\bottomrule
\end{tabular}

\label{tab:tile_ablation}
\end{table}

Although the 50-tile setting achieved a slightly higher mean QWK, 100 tiles showed the lowest standard deviation across repeated runs and was retained as the main setting. These results suggest that ProsMAE is not highly sensitive to the exact number of sampled tiles within the evaluated range.

\section{Conclusion}
This paper presented ProsMAE for multi-source MAE pretraining and ProsCLS for ISUP grade classification. The proposed pipeline is designed as a low-compute and deployment-friendly framework, using only 5000 MAE pretraining steps, a frozen encoder, mean-pooled WSI features, and a lightweight linear probe for classification. Under the primary disjoint PANDA split, multi-source pretraining improved mean validation QWK over the vanilla MAE baseline. Because downstream evaluation is performed on a single PANDA cohort and primary split, robustness across external cohorts cannot yet be claimed. Future work will include repeated validation on independent prostate cancer cohorts to verify generalization.

\section*{Acknowledgment}
This work was supported by the IITP grant (IITP-2023-RS-2023-00256081) funded by MSIT, Korea, and the ANCHOR program (2026-ANCHOR-01-110) funded by the Ministry of Education and the Seoul Metropolitan Government, Republic of Korea.


\bibliographystyle{IEEEtran}
\bibliography{ref}

\end{document}